\documentclass[preprint,12pt]{elsarticle}

\usepackage{amssymb}
\usepackage{threeparttable}
\usepackage{multirow}
\usepackage{amsmath}
\usepackage{subfigure}
\usepackage{diagbox}
\usepackage{hyperref}

\usepackage{lineno}

\usepackage{xcolor}

\journal{Automation in Construction}

\begin{document}

\begin{frontmatter}



\title{Attention-Enhanced Co-Interactive Fusion Network (AECIF-Net) for Automated Structural Condition Assessment in Visual Inspection}


\author[inst1]{Chenyu Zhang}

\affiliation[inst1]{organization={Department of Civil Engineering, Stony Brook University},
            addressline={2427 Computer Science}, 
            city={Stony Brook},
            postcode={11794}, 
            state={New York},
            country={United States}}

\author[inst2]{Zhaozheng Yin}

\affiliation[inst2]{organization={Department of Computer Science, Department of Biomedical Informatics, and AI Institute, Stony Brook University},
            addressline={2313B Computer Science Building}, 
            city={Stony Brook},
            postcode={11794}, 
            state={New York},
            country={United States}}

\author[inst1]{Ruwen Qin}

\begin{abstract}
Efficiently monitoring the condition of civil infrastructure requires automating the structural condition assessment in visual inspection. This paper proposes an Attention-Enhanced Co-Interactive Fusion Network (AECIF-Net) for automatic structural condition assessment in visual bridge inspection. AECIF-Net can simultaneously parse structural elements and segment surface defects on the elements in inspection images. It integrates two task-specific relearning subnets to extract task-specific features from an overall feature embedding. A co-interactive feature fusion module further captures the spatial correlation and facilitates information sharing between tasks. Experimental results demonstrate that the proposed AECIF-Net outperforms the current state-of-the-art approaches, achieving promising performance with 92.11\% mIoU for element segmentation and 87.16\% mIoU for corrosion segmentation on the test set of the new benchmark dataset Steel Bridge Condition Inspection Visual (SBCIV). An ablation study verifies the merits of the designs for AECIF-Net, and a case study demonstrates its capability to automate structural condition assessment.

\end{abstract}


\end{frontmatter}


\section{Introduction}
\label{sec:intro}




Visual inspection, a crucial component of structural health monitoring (SHM), is performed periodically to evaluate the condition of infrastructure~\citep{graybeal2002visual}. However, traditional manual inspections have inherent limitations. Desires for time-cost efficiency, reliability, and safety have driven a growing interest in automating visual inspection with cutting-edge technologies like robotics and artificial intelligence~\citep{drones6110355}. Unmanned aerial vehicles (UAVs), equipped with one or multiple types of non-destructive evaluation sensors, have gained popularity for capturing inspection videos and images of infrastructure~\citep{OUTAY2020116}. Maximizing the potential of robotic inspection platforms and the automation process necessitates the employment of efficient and reliable techniques for inspection image analysis. Deep convolutional neural networks (DCNNs), in particular, have shown tremendous potential for analyzing images and extracting vital information about the inspected structures, inspiring researchers to investigate their applications in SHM~\citep{spencer2019advances}. For example, a drone with mounted RGB cameras can quickly assess the condition of a bridge at the inspection site and narrow down to spots where other high-resolution yet time-consuming diagnostic sensors should be used to collect detailed information, such as infrared sensors, ground-penetrating radar, ultrasound scanning, and others.

According to the infrastructure inspection manuals and standards~\citep{inspection, aashto2019manual,hartle2002bridge,astm2015standard}, it is necessary to associate structural elements with the severity of defects developed in the elements to evaluate the condition of individual elements, which builds the foundation for assessing the condition of the overall structure. That is, it is required to not only recognize and localize key structural elements and defects in images captured by inspection robots, but also spatially associate them. This capability will offer a reference for prioritizing subsequent structural condition assessment that is usually more expensive.

\begin{figure*}[htbp]
    \centering
    \includegraphics[width=\textwidth]{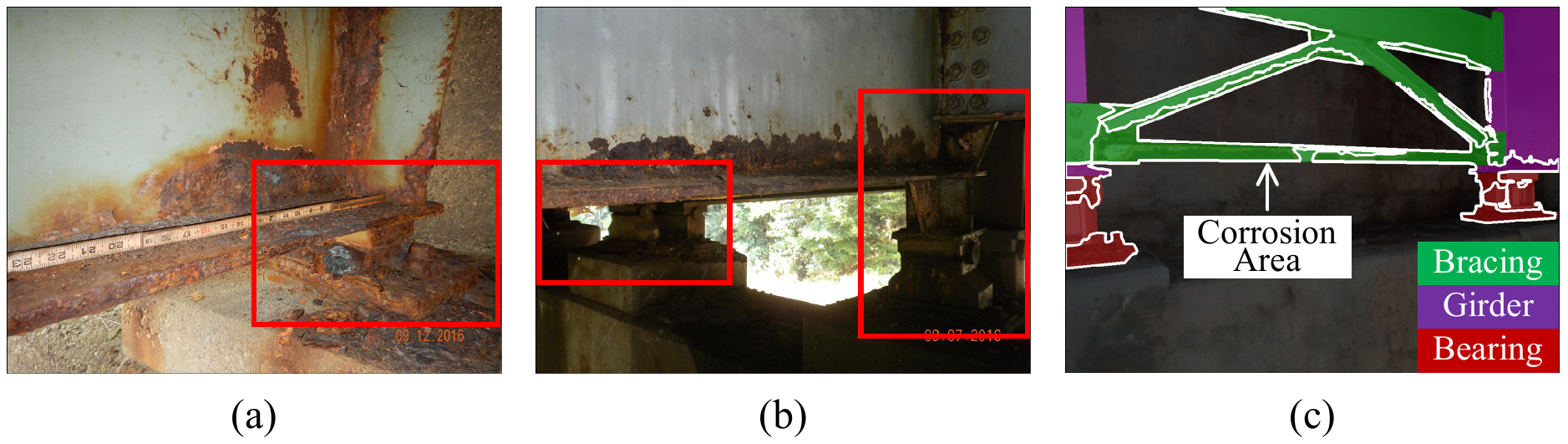}
    \caption{Challenges in deep learning-based visual inspection: (a) potential confusion due to surface defects, where a rusted girder might be mistakenly identified as a bearing; (b) issues from surface inhomogeneity, shadows, and poor lighting affecting defect assessment; (c) overlooked spatial correlation between element segmentation and defect segmentation tasks. (Origin images courtesy of~\citep{bianchi2021coco}).}
    \label{fig:chall}
\end{figure*}

Researchers have made progress in identifying or segmenting structural elements and defects using DCNNs~\citep{https://doi.org/10.1111/mice.12505,karim2022semi,BIANCHI2022104299}. However, most studies were dedicated to addressing one task, leading to three challenges in deep learning-based visual inspection. Firstly, the appearance of structural elements may be inaccurately recognized due to the presence of surface defects. For example, Fig.~\ref{fig:chall}(a) shows a rusted girder section and a below-bearing share similar surface defects. The girder's rusted or flaking portion might be mistakenly identified as part of the bearing due to the similarities in appearance caused by the defects. Secondly, surface inhomogeneity, shadows, and poor lighting conditions, as shown in Fig.~\ref{fig:chall}(b), continue to pose challenges for reliably assessing defects on surfaces of structural elements. Lastly, the spatial correlation between element segmentation and defect segmentation tasks, as shown in Fig.~\ref{fig:chall}(c), has been overlooked, leading to unreasonable predictions. For example, the presence of steel corrosion in the background area is apparently a wrong prediction result because it is impossible in reality. Several attempts have explored solutions to these challenges, mainly using multi-task learning (MTL) methods~\citep{hoskere2020madnet, https://doi.org/10.1002/stc.3128, doi:10.1177/03611981231155418}. MTL is an efficient approach that learns to perform the two related tasks simultaneously with a unified model~\citep{zhang2021survey}. While those studies have laid a solid foundation for visual assessment of the structure's condition, several research needs must be addressed further to improve the technology readiness of image analysis for automated visual inspection.

MTL can adopt a powerful deep encoder that generates a deep feature embedding to represent each input inspection image~\citep{zhang2021survey}. There are many choices of deep encoders for semantic segmentation. The guidance for choosing one deep encoder that is well-suited for the tasks of this paper has not been available. Moreover, the overall embedding encompasses information related to both structural elements and surface defects. While a simple method named feature projection~\citep{doi:10.1177/03611981231155418} has been developed to decouple task-specific features intertwined in the overall embedding, a more advanced method to extract task-specific features called feature relearning has not been explored yet. Last but not least, how to leverage the spatial correlation between structural elements and defects to let one task benefit from the other task and vice versa is still an unsolved question. A cross-talk method~\citep{doi:10.1177/03611981231155418} was created for this purpose. However, that design does not explicitly integrate the physical meaning of spatial information that one task can provide to another. The spatial attention mechanism, which focuses on spatial attributes such as shape and boundaries within an image, is critical in image analysis. Upon obtaining the spatial attention maps and task-specific features, the design of a suitable network architecture becomes pivotal. Such an architecture should facilitate efficient communication and information exchange across tasks. Incorporating this diverse information can enhance a model's understanding of the complex interrelations within the data, leading to more robust feature representations and improved semantic segmentation outcomes.

In addressing the above-discussed technical needs, this paper has the following contributions:

\begin{itemize}
\item A new MTL model, named Attention-Enhanced Co-Interactive Fusion Network (AECIF-Net), is introduced. It has a share-split-interaction pipeline composed of a shared high-resolution deep encoder, two task-specific relearning subnets, and a co-interactive feature fusion module.

\item A new dataset, named Steel Bridge Condition Inspection Visual (SBCIV) dataset, is developed to support the development and evaluation of MTL models for automating the bridge element inspection.

\item A comprehensive study employing both numerical experiments and qualitative evaluation is conducted to verify the strengths of AECIF-Net and reveal reasons for achieving satisfying performance.
\end{itemize}

\begin{figure*}[htbp]
    \centering
    \includegraphics[width=\textwidth]{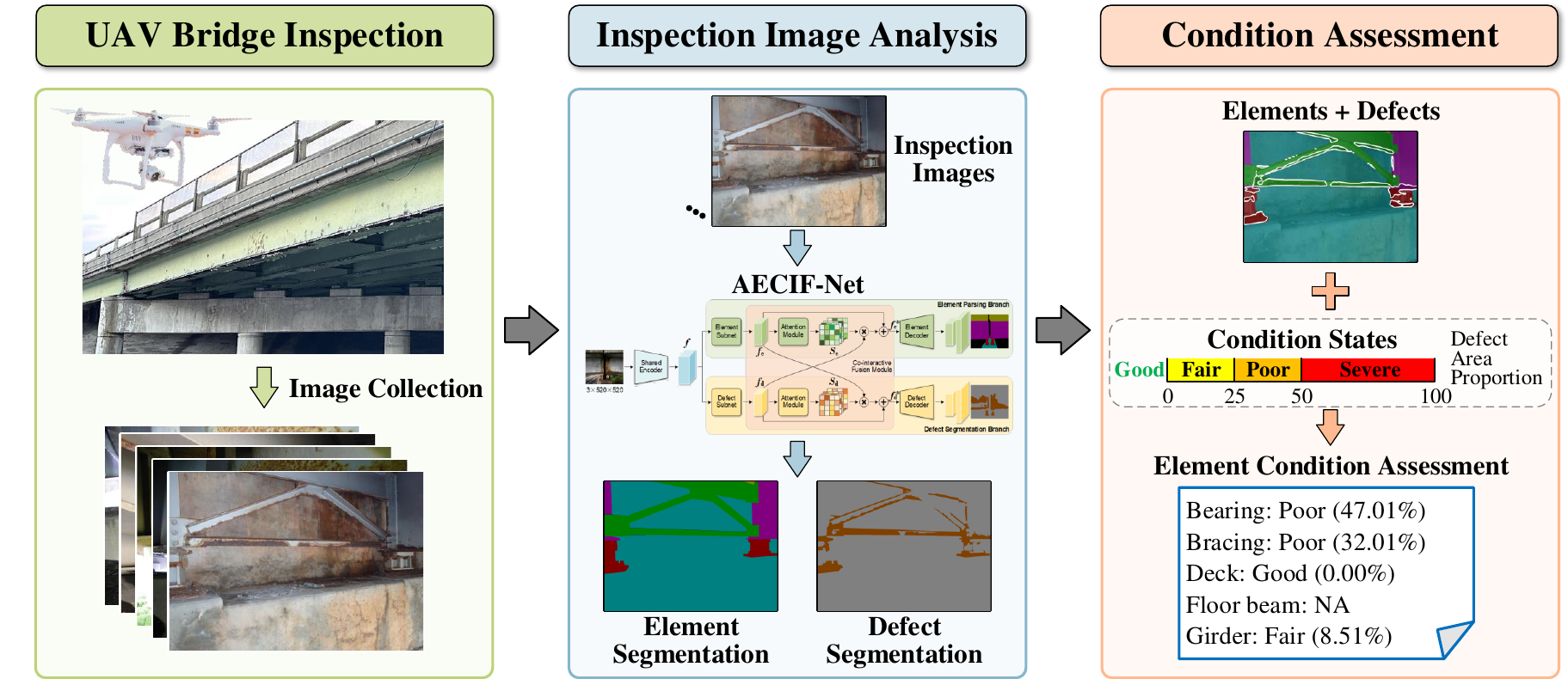}
    \caption{Framework of the automated visual bridge inspection using UAVs, where AECIF-Net analyzes collected images to segment structural elements and defects, leads to a comprehensive structural condition assessment. (Example images courtesy of Shengqian Zheng and~\citep{bianchi2021coco}.)}
    \label{fig:fram}
\end{figure*}

Based on the proposed AECIF-Net, the automatic structural condition assessment framework in visual bridge inspection is shown in Fig.~\ref{fig:fram}.

The remainder of the paper is organized as follows. The next section is a summary of related work. Then, Section~\ref{sec:network} presents the design of the AECIF-Net, followed by details of executing the model. Section~\ref{sec:exp} discusses results from the experimental studies for evaluating AECIF-Net, and Section~\ref{sec:case} further presents an assessment case study. In the end, Section~\ref{sec:con} summarizes insights gained from this study and suggestions for important future work. The code and dataset used in this study are available at \url{https://github.com/itschenyu/AECIF-Net}.

\section{Related work}
\label{sec:work}

This paper is built on studies that contributed to structural condition assessment in visual inspection, either directly or indirectly. The related literature is summarized below.

\subsection{DCNN-based defect segmentation}

An intensively studied topic related to the visual inspection of infrastructures is defect detection, which is about finding structural surface defects or damage in inspection images or videos~\citep{YANG2022129226}. The majority of current research efforts are centered on DCNN-based defect segmentation, where each pixel of an inspection image is classified as defect or non-defect~\citep{HU2021103973}. Segmentation can provide pixel-level position information of defects, resulting in superior accuracy compared to object detection methods~\citep{ZHOU2023104678}. Deep feature extractors, such as DCNNs, are employed due to their ability to facilitate automated representation learning and embed rich information, ultimately capturing complex real-world data features through multi-level feature abstraction~\citep{khan2020survey}.

DCNN-based crack segmentation methods have shown considerable success in detecting and analyzing defects across various civil structures, including buildings, bridges, tunnels, and roads~\citep{ALI2022103989}. \citet{DUNG201952} utilized a Fully Convolutional Network (FCN)~\citep{long2015fully} for crack element marking, while \citet{JI2020103176} aimed at accurately quantifying cracks by training an atrous convolution-based DeepLabv3+ model~\citep{chen2018encoder}. \citet{MEI2020103018} focused on crack connectivity through a densely connected DCNN architecture. \citet{LIU2022126265} incorporated visual explanations into a U-Net~\citep{ronneberger2015u} based model to highlight crack semantics.

Recently, DCNNs have been utilized for the detection and segmentation of corrosion in steel structures. \citet{doi:10.1177/1475921717737051} demonstrated that DCNNs surpass conventional vision-based corrosion detection methods that rely on texture and color analysis using a basic multi-layer perceptron network. \citet{rahman2021semantic} introduced a corrosion assessment approach that applied DeepLab~\citep{7913730} to infrastructure inspection images. \citet{han2021recognition} developed a two-stage corrosion location method by integrating Feature Pyramid Network (FPN)~\citep{lin2017feature} and Path Aggregation Network (PANet)~\citep{liu2018path} to identify corrosion areas on structural surfaces. \citet{jiang2023automatic} proposed an enhanced U-net, Fusion-Attention-U-net (FAU-net), which incorporated a fusion module and an attention module within the U-net for segmenting three types of corrosion-related damage within dim steel box girders. \citet{KATSAMENIS2022104182} applied U-Net for the automated simultaneous detection and localization of corrosion and rust grade recognition from inspection images of metal structures. These studies have laid a solid methodological groundwork for identifying defects in structural elements.

\subsection{DCNN-based structural element segmentation}

Structural element inspection requires associating elements with defects developed on them. A stream of recent studies was motivated to focus on structural element segmentation in inspection images using various DCNN-based methods. This segmentation process aims at identifying structural elements within these images, typically achieved through object detection or segmentation within the given scene. Accurate identification of critical elements enables a thorough and precise evaluation of the infrastructure's overall condition, considering factors like defect shape, size, location, and compliance with established standards. \citet{https://doi.org/10.1111/mice.12363} developed a hierarchical framework applicable to a range of classification tasks, including recognizing building component types of damage states. \citet{doi:10.1177/1475921718765419} detected damage in welded joints on truss structures by extracting and classifying target image areas. \citet{https://doi.org/10.1111/mice.12505} presented an FCN-based method for bridge component recognition. \citet{CZERNIAWSKI2020101131} developed a DeepLab-based model incorporating RGB-D (color and depth) images for component segmentation in thirteen buildings. \citet{WANG2023110028} proposed an enhanced U-Net model with a novel geometric consistency loss for geometry-informed structural component segmentation of post-earthquake buildings. \citet{karim2022semi} transferred a pre-trained Mask R-CNN~\citep{he2017mask} to the task of bridge elements segmentation and created a semi-supervised self-training method to refine the transferred network iteratively. Through a comparative analysis of the state-of-the-art semantic segmentation networks, \citet{zhang2022adeep} revealed the aptitude of High-Resolution Network (HRNet)~\citep{wang2020deep} in efficiently extracting deep features for segmenting various structural elements in bridge inspection images. Furthermore, the study investigated factors that impact the network's performance, such as transfer learning, the size of the training set, data augmentation techniques, and the role of class weights.

Although impressive, the above methods exhibit some limitations in identifying and capturing the highly irregular and significantly deteriorated elements from inspection images. Furthermore, these approaches have yet to explore feature fusion's potential to enhance element segmentation capability. Notably, compared to the extensively studied defect segmentation, structural element segmentation remains in a relatively nascent stage of development.

\subsection{MTL in visual structural assessment}

MTL has proven effective in various civil engineering applications, such as SHM data reconstruction~\citep{doi:10.1177/1475921718794953}, bridge damage diagnosis~\citep{doi:10.1177/14759217221081159}, landslide evolution state prediction~\citep{SUN2022107884}, and more. The prevalent method for achieving MTL is to share the feature extractor and branch downstream tasks for respective predictions~\cite{vandenhende2021multi}. This shared feature extractor learns a common representation for all tasks, significantly reducing the risk of over-fitting and enhancing generalization~\citep{zhang2021survey}. However, one task can easily dominate others in this conventional approach, which negatively impacts the overall performance.

\citet{hoskere2020madnet} introduced MaDnet, a DCNN comprising a shared feature extractor and multiple semantic segmentation pathways to identify material and damage types. This framework indicates that one segmentation task can provide contextual information for another. \citet{https://doi.org/10.1002/stc.3128} developed the MT-HRNet that employs the HRNetV2-W18 backbone and two segmentation heads for element recognition and damage identification in synthetic bridge images. To further address the challenges of task domination and improve task-specific feature sharing, \citet{doi:10.1177/03611981231155418} proposed the MTL-D and MTL-I models that both can simultaneously segment bridge elements and surface corrosion. These models project the shared features for respective tasks and utilize the cross-talk feature sharing between tasks to enhance performance and prevent the dominance of a single task. 

Despite remarkable advancements in this direction, most studies remain at the stage of Naive MTL models. While attempts were observed to exchange information among different tasks, they have yet to explicitly model the spatial association between elements and defects. The effective utilization of this spatial relationship will mitigate the task domination issue and enhance the overall performance, which remains a further exploration.

\section{AECIF-Net}
\label{sec:network}

\begin{figure*}[htbp]
    \centering
    \includegraphics[width=\textwidth]{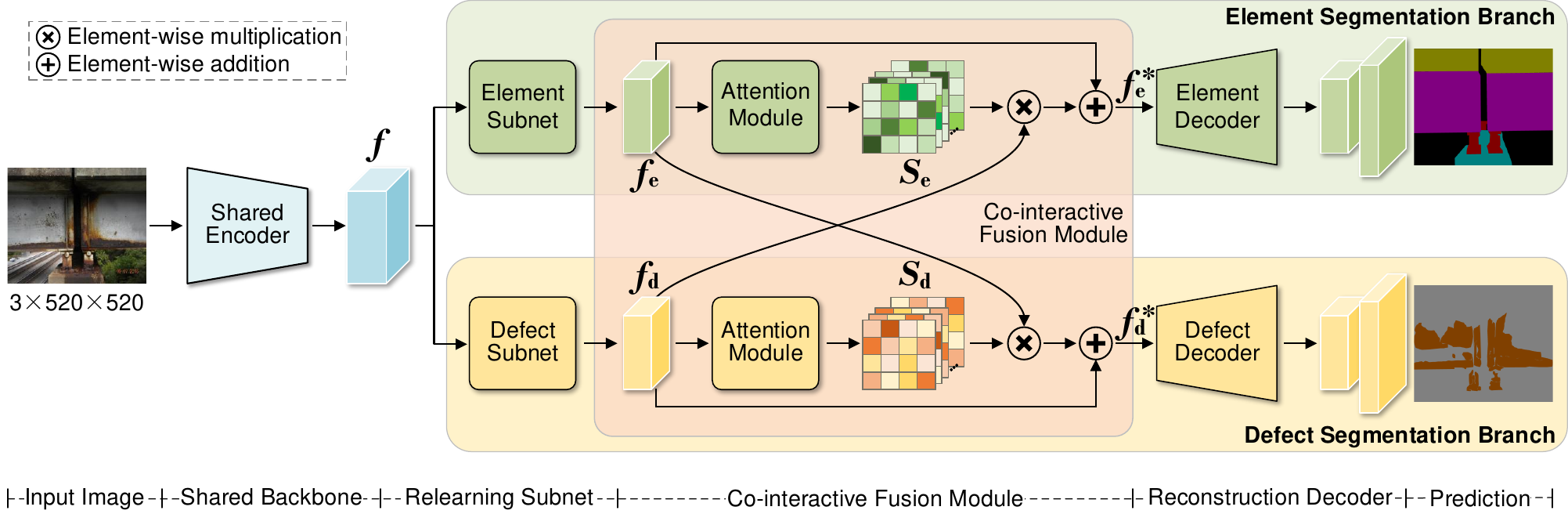}
    \caption{Architecture of the AECIF-Net, which features a share-split-interaction pipeline composed of a shared high-resolution deep encoder, two task-specific relearning subnets, and a co-interactive feature fusion module.}
    \label{fig:MTL_architecture}
\end{figure*}

In filling the above-discussed gaps, this paper introduces an Attention-Enhanced Co-Interactive Fusion Network (AECIF-Net), shown in Fig.~\ref{fig:MTL_architecture}. AECIF-Net analyzes each input RGB image to parse the bridge in the image into structural elements and segment the surface defects on the elements. The backbone of AECIF-Net encodes the input image as an overall feature embedding. Then, two relearning subnets respectively extract element- and defect-specific feature maps from the encoded overall embedding. After that, a co-interactive module generates spatial attention maps to guide the feature fusion for performing the two downstream tasks. Finally, two reconstruction decoders respectively perform the pixel-level classification for element segmentation and defect segmentation. Details of the AECIF-Net's key components are delineated below.

\subsection{Shared encoder}

AECIF-Net employs HRNetV2-W48~\citep{wang2020deep} as its shared encoder, attributed to its capability to produce high-resolution feature maps through parallel high-to-low-resolution convolutions. Before entering the shared encoder, an inspection image captured by the robot-mounted RGB camera is reshaped to be the size $3\times 520\times 520$. The shared encoder serves as a feature extractor to capture an overall feature embedding, $\pmb{f}(\in\mathbb{R}^{720\times 120\times 120})$, from the input image. The overall feature embedding integrates multi-dimensional information, encompassing both structural elements and surface defects. Using the shared encoder improves computational efficiency, ensuring the extraction of comprehensive, effective, and non-redundant feature embedding.

\subsection{Task-specific feature relearning subnets}

$\Omega=\{\rm e, \rm d\}$ is the index set of tasks, where $\rm e$ and $\rm d$ stand for element segmentation and defect segmentation tasks, respectively. These two tasks concentrate on distinct characteristics and minutiae. Therefore, two task-specific relearning subnets follow the shared encoder. These subnets refine the overall feature embedding from the shared encoder to further extract task-specific feature maps with lower dimensions, $\pmb{f}_i$ ($\in\mathbb{R}^{512\times 120\times 120})$, for $i\in \Omega$. They enhance the precision in feature extraction and optimize computational performance, which is crucial for reliable bridge condition assessments.

Each relearning subnet consists of a convolutional layer (COV), a batch normalization layer (BN), and the rectified linear unit (ReLU) activation function in sequence:
\begin{equation}
    \pmb{f}_{i} = \mathrm{ReLU(BN(COV}\left(\pmb{f}; \boldsymbol{\theta}_{{\rm rln},i}))\right),\; \forall i\in\Omega.
    \label{eq:relearning}
\end{equation}

The convolutional operation in Eq.~(\ref{eq:relearning}) uses a kernel size of 3, a stride of 1, padding of 1, and 512 output channels. $\pmb{\theta}_{{\rm rln},i}$ are learnable parameters of the convolutional layer in the relearning (rln) subnet for task $i$, with a total of 3.32 million parameters in each subnet.

\subsection{Co-interactive fusion module}

The co-interactive fusion module is an essential component in the AECIF-Net, acting as a bridge between the element and defect segmentation branches. This is conceived with the understanding that structural elements and defects often exhibit spatial correlations in bridge inspection images. Recognizing that these tasks are not mutually exclusive but rather complementary, the module is strategically designed to enable information exchange and knowledge transfer between tasks. In this way, the performance of each task is enhanced. More technically, the module employs an additive fusion, wherein the task-specific feature map for one task is enriched with additional spatial information from the other:
\begin{equation}
    \pmb{f}_{i}^{*} = \pmb{f}_{i} \oplus \left(\pmb{S}_{i} \otimes \pmb{f}_{j}\right),\; \forall i, j\in\Omega\, \text{and}\, i\neq j
    \label{eq:fusion}
\end{equation}
Here, $\otimes$ represents the element-wise multiplication, $\oplus$ denotes the element-wise addition, $\pmb{S}_{i}$ ($\in\mathbb{R}^{512\times 120\times 120})$ is the spatial attention mask for guiding the feature fusion for task $i$, and $\pmb{f}_i^\ast$ is the resulting spatial attention-enhanced feature map for the task.


As specified in Eq.~(\ref{eq:fusion}), the spatial attention mask for one task is used to assign scores that modify the other task's feature map during this fusion process. The mechanism thus works to fine-tune each task's features based on the complementary strengths of the other, contributing to the model's performance improvement. These scores are learned by a convolutional layer (with a kernel size of 3, a stride of 1, and padding of 1) and normalized using the Sigmoid function:  
\begin{equation}
    \pmb{S}_i = \mathrm{Sigmoid}(\mathrm{COV}(\pmb{f}_i;\pmb{\theta}_{{\rm att},i})),\;\forall i\in\Omega
\end{equation}
where $\boldsymbol{\theta}_{{\rm att},i}$ are the learnable parameters of the attention (att) module for task $i$, with a total of 2.36 million parameters in each attention module.

\subsection{Reconstruction decoder}

In leaving the co-interactive fusion module, the spatial attention-enhanced feature map for any task $i$, $\pmb{f}_i^\ast$, flows into the task's reconstruction decoder. This component plays a pivotal role in converting the abstract feature representation into meaningful segmentation predictions. First, the segmentation head (SH) in the decoder performs the pixel-level classification, which is composed of a convolutional layer (with a kernel size of 1, a stride of 1, and 512 output channels), a batch normalization layer, a ReLU activation function, and another convolutional layer (with a kernel size of 1, a stride of 1, and $N_i$ output channels). Then, the obtained segmentation map is upsampled (UP) using the bilinear interpolation to give the pixel-level prediction scores for all classes, $\widehat{\pmb{y}}_i$ ($\in \mathbb{R}^{N_i\times 520\times 520}$). That is, 
\begin{equation}
    \widehat{\pmb{y}}_{i}=\mathrm{UP}(\mathrm{SH}(\pmb{f}^\ast_i;\pmb{\theta}_{{\rm sh},i})),\; \forall i\in\Omega
\end{equation}
where $\pmb{\theta}_{{\rm sh},i}$ are learnable parameters of the convolutional layers in the segmentation head of task $i$, with 0.27 million parameters in each segmentation head.

\subsection{Loss function}

Learnable parameters of the proposed AECIF-Net are determined through model training that minimizes a differentiable loss function through backpropagation. The loss function is an aggregated measure of the pixel-level dissimilarity between the ground truth and prediction on a training dataset.

Images in the training set are indexed by $k$, $\pmb{y}_i (k)$ ($\in R^{N_i\times 520\times 520}$) represents the one-hot encoding of the image's pixel-level ground truth associated with task $i$, and $\widehat{\pmb{y}}_i(k)$ ($\in R^{N_i\times 520\times 520}$) denotes the pixel-level prediction. The cross-entropy loss of AECIF-Net in performing task $i$ is
\begin{equation}
  \mathcal{L}_i = - \sum_k <\pmb{y}_i(k), \log\widehat{\pmb{y}}_i(k)>,\; \forall i\in\Omega
  \label{eq:losses}
\end{equation}
where $<,>$ is the operation to obtain the Frobenius inner product on two tensors, which performs the element-wise product of the two input tensors to become one in the same size and then sums up all the elements of the resulting tensor.

The loss functions defined for the individual tasks need to be integrated as a total loss function so that AECIF-Net learns the two tasks at once. This paper employs a straightforward yet efficient weighting scheme, known as Dynamic Weight Average (DWA)~\citep{liu2019end}, to adaptively balance the individual loss functions during training. Given that $t$ is the index of training epochs and $\mathcal{L}_{i,t}$ is the loss function of task $i$ calculated using Eq.~(\ref{eq:losses}) at epoch $t$, the relative loss descending rates of the two tasks in the last training epoch are:
\begin{equation}
    \pmb{w}_{t-1}=\left[\frac{\mathcal{L}_{{\rm e},t-1}}{\mathcal{L}_{{\rm e},t-2}}, \frac{\mathcal{L}_{{\rm d},t-1}}{\mathcal{L}_{{\rm d},t-2}}\right]
    \label{eq:lossdecendingrate}
\end{equation}

These rates are references for assigning weights to the individual loss functions at the current epoch. $\pmb{w}_t$ is initialized as $[1,1]$ at $t=1, 2$.
The weights $\lambda_{{\rm e},t}$ and $\lambda_{{\rm d},t}$ for aggregating the individual loss functions are obtained by applying the Softmax operation to $\pmb{w}_{t-1}$, 
\begin{equation}
\pmb{\lambda}_t=2\cdot\mathrm{Softmax}[\pmb{w}_{t-1}/\tau]
\label{eq:weights}
\end{equation}
where $\pmb{\lambda}_t=[\lambda_{{\rm e},t}, \lambda_{{\rm d},t}]$ is the vector of weights at $t$, and $\tau$ is a temperature parameter for controlling the softness of this weighting scheme, chosen as 2 in this study. 
Multiplying the outcomes of the Softmax function by a scaling factor of two ensures $\sum_{i \in \Omega} \lambda_{i,t}=2$. This setting is consistent with a default scenario where, without any weighting method, every task receives a weight of 1.

The total loss function, $\mathcal{L}_{\rm tot}$, is attained by aggregating the individual training loss functions $\mathcal{L}_{{\rm e},t}$ and $\mathcal{L}_{{\rm d},t}$ using the weights calculated in Eq.~(\ref{eq:weights}):
\begin{equation}
    \begin{aligned}
        &\mathcal{L}_{\rm tot} = \lambda_{{\rm e},t}\mathcal{L}_{{\rm e},t}+\lambda_{{\rm d},t}\mathcal{L}_{{\rm d},t} \\
    \end{aligned}
\end{equation}

With the DWA scheme, the task with the larger relative loss descending rate is the slower learner. As such, a recent slower-learning task receives a larger weight ($>1$), while the other faster one obtains a smaller weight ($<1$), in the next training epoch.

\section{Experimental setup}
\label{sec:setup}
This section discusses the details of the dataset development, model implementation, and evaluation metrics.

\subsection{Dataset and data augmentation}

Lacking publicly available datasets with the annotations for developing the proposed MTL model, this study created a new dataset, called the Steel Bridge Condition Inspection Visual (SBCIV) dataset, to address the pressing need for appropriate data. Distinct from existing datasets, the SBCIV dataset offers detailed pixel-level annotations from genuine bridge inspection images, focusing on structural element and surface defect segmentation. This unique dataset not only advances the development and validation of MTL models, but also establishes a novel benchmark for research in SHM and visual inspection.
\subsubsection{Data collection}

The SBCIV dataset comprises 440 high-resolution images procured from two publicly available datasets: the Common Objects in Context for Bridge Inspection (COCO-Bridge) dataset~\citep{bianchi2021coco} and Corrosion Condition State Semantic Segmentation dataset~\citep{Bianchi2021}.

The COCO-Bridge dataset is an image-based dataset assisting UAVs in element identification. It consists of 774 images, focusing on identifying specific parts of bridges or structural details to make autonomous decisions during flight. The Corrosion Condition State Semantic Segmentation dataset has 440 annotated images. It provides four corrosion class categories: good, fair, poor, and severe. The annotations are aligned with the corrosion condition state guidelines in the Manual for Bridge Element Inspection\citep{aashto2019manual} and the Bridge Inspector's Reference Manual~\citep{hartle2002bridge}. Both datasets are created from the Virginia Department of Transportation's Bridge Inspection Reports and have been acknowledged by the department for their practical use in bridge condition assessment. Besides, these datasets are also gaining popularity in SHM and visual inspection research (e.g., ~\citep{doi:10.1061/(ASCE)CP.1943-5487.0001045, doi:10.1177/14759217221083647, eltouny2023highresolution, doi:10.1177/03611981231155418, LIU2023107014, WANG2023105085}).

\subsubsection{Data annotation}

\begin{figure*}[htbp]
    \centering
    \includegraphics[width=0.8\textwidth]{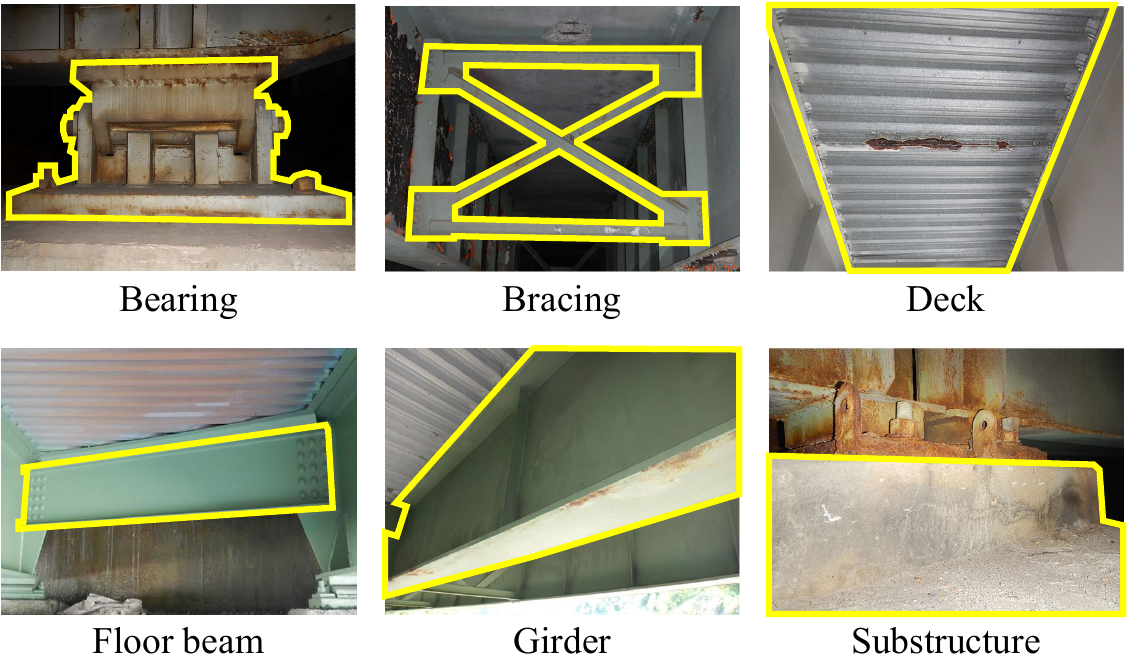}
    \caption{Illustration of bridge element classes that collectively define the structural area within inspection images. (Base images courtesy of \citep{bianchi2021coco} and \citep{Bianchi2021}.)}
    \label{fig:data}
\end{figure*}

Annotations of the SBCIV dataset provide pixel-level labels for critical structural elements and surface defects of steel plate girder bridges. As illustrated in Fig.~\ref{fig:data}, the dataset covers six predominant structural elements: bearing, bracing, deck, floor beam, girder, and substructure. These elements collectively define and parse the entire structural area within the inspection images.

The pixel-level defect annotation for SBCIV is binary: corrosion and non-corrosion. Images from the Corrosion Condition State Semantic Segmentation dataset were initially labeled as four classes, which were converted into binary labels in this study. Images selected from the COCO-Bridge dataset had no defect annotation. This study semantically labeled defects in those images and ensured the annotation consistency between the two data sources.

To ensure the accuracy and reliability of data annotation, the LabelMe~\citep{russell2008labelme} labeling tool, known for its precision and user-friendly interface, was employed during the annotation process. To maintain rigorous consistency and precision, the Bridge Inspector's Reference Manual and the corrosion condition state guidelines outlined in the Manual for Bridge Element Inspection were strictly followed.

\subsubsection{Dataset split and class distribution}

For evaluating the performance of AECIF-Net, a set of 100 images was reserved as the test set. The remaining 340 images are further split into the training and validation sets according to a 9:1 ratio. The 306 training images are used for comprehensively training the model, whereas the 34 validation images are for performance monitoring and hyperparameter optimization in the training stage. Separating data for training, validation, and testing is a model development strategy for preventing over-fitting and improving generalization. The data distribution chosen by this study ensures adequate data for model development and independent evaluations in both training and inference.

\begin{figure*}[htbp]
    \centering
    \subfigure[]{
    \label{dist1}
    \includegraphics[width=0.48\textwidth]{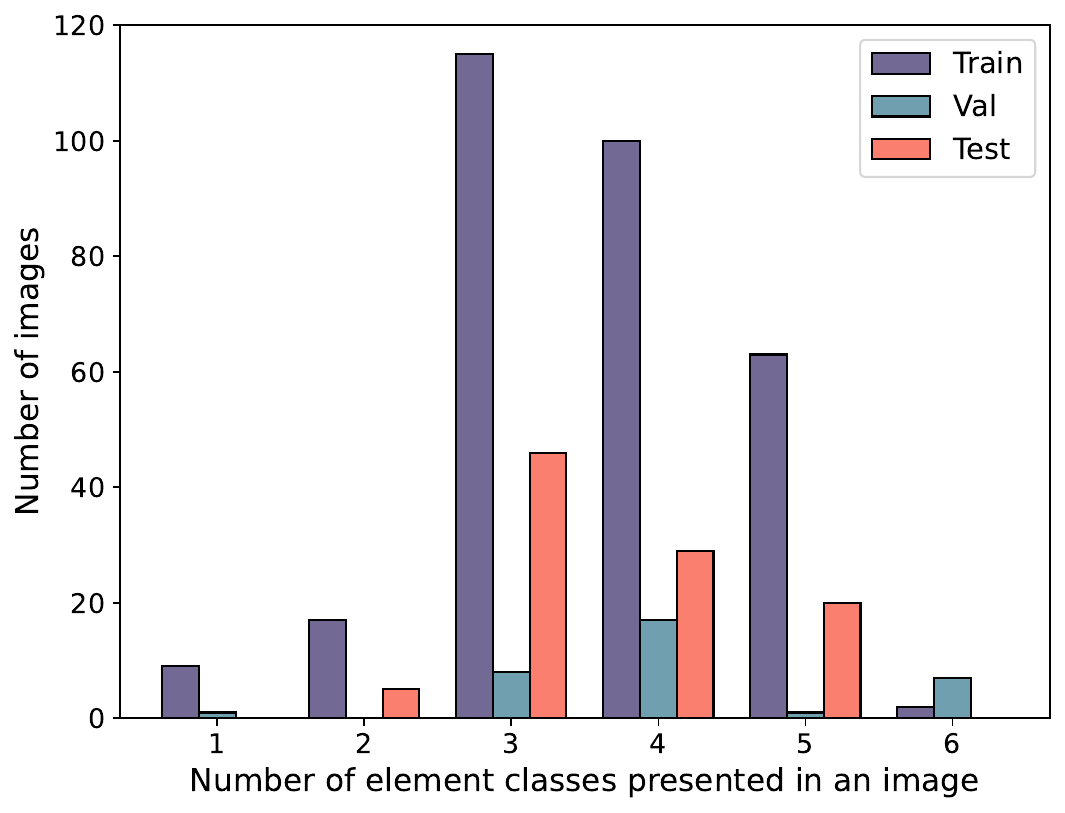}}
    \subfigure[]{
    \label{dist2}
    \includegraphics[width=0.48\textwidth]{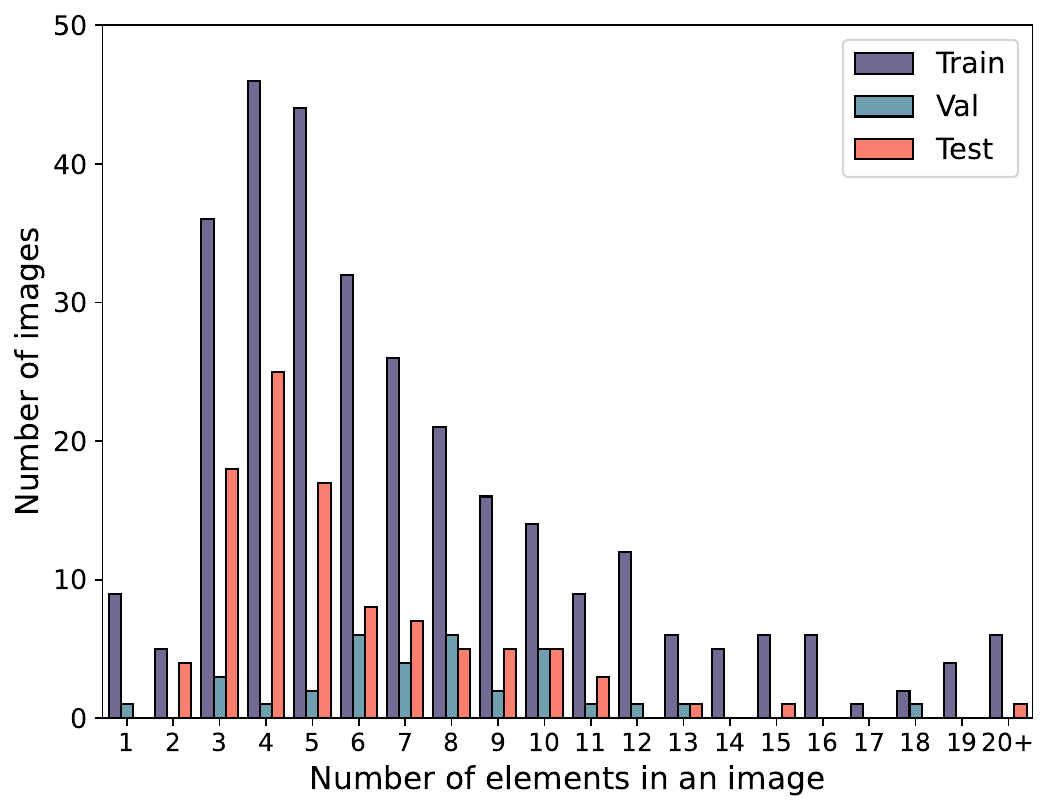}}
    \caption{Data distributions of the SBCIV dataset, split by training, validation, and test: (a) by the number of element classes in an image, (b) by the number of elements in an image. The dataset contains a wide range of different examples that are important for teaching models to handle various situations.}
    \label{fig:dist}
\end{figure*}

The number of element classes and the number of elements varies from one image to another. Fig.~\ref{fig:dist} shows the frequency distributions of images by these two variables, split by training, validation, and test. Fig.~\ref{fig:dist} (a) indicates that most images contain 3$\sim$5 element classes. The number of elements per image spans from 1 to 20+. Such diversity prepares models for both simple and complex assessment scenarios. The data distribution among training, validation, and test sets is appropriate, ensuring comprehensive model training, accurate optimization, and unbiased evaluation.


Deep learning models usually require a large amount of data to be trained effectively. Yet, creating a large dataset with the required annotation for the problem of study is expensive. The issue of small data can be partially addressed by data augmentation that aims to not only increase the data quantity but also cover situations that are not in the original dataset but could occur in the real world. To achieve this, five image data augmentation methods were employed in this paper, including random scale transformation, random rotations between $\pm10^{\circ}$, random horizontal flipping, random image intensity noise using a $5\times5$ Gaussian kernel, and HSV augment that randomly adjusts hue (H), saturation (S), and value (V) of images.

\subsection{Implementation details}

The AECIF-Net was built based on the PyTorch 1.10.0 library and trained on a server with an Nvidia Tesla V100 GPU (32 GB memory). The Adam optimizer with an initial learning rate of 5e-4 and a minimum learning rate of 5e-6 was utilized for training the model. A cosine learning rate scheduler was used to adjust the learning rate during training. Owing to the limited dataset size in this study, transfer learning was applied, where the backbone of AECIF-Net was initialized with weights pre-trained on the Cityscapes dataset~\citep{cordts2016cityscapes}. The model was fine-tuned for 150 epochs with a batch size of 8. For computational efficiency, all images were resized from the original size to $520\times 520$ pixels. The model achieving the lowest loss on the validation set was saved as the final model.

The AECIF-Net was evaluated through comparative studies that measure its performance against related models and state-of-the-art summarized below.
\begin{itemize}
    \item {\it Single-task models}: Two models, that use HRNetV2-W48 as the backbone, were trained separately using the hyperparameters mentioned above. They perform the structural element segmentation and defect segmentation tasks, respectively. The two single-task models serve as the baseline for assessing MTL models in this paper.
    \item {\it Variants of AECIF-Net}: Three variants of AECIF-Net were trained to evaluate the designs of AECIF-Net. The Naive MTL model drops both the task-specific relearning (TR) subnets and the co-interactive fusion (CF) module from the AECIF-Net. The AECIF-Net without TR is the model that drops only the task-specific relearning subnets, whereas the AECIF-Net without CF drops only the co-interactive fusion module.
    \item {\it State-of-the-art models}: The recently developed MTL models, including MaDnet~\citep{hoskere2020madnet}, MT-HRNet~\citep{https://doi.org/10.1002/stc.3128}, and MTL-I~\citep{doi:10.1177/03611981231155418}, have shown state-of-the-art performance in segmenting both bridge elements and surface defects. AECIF-Net was compared to these models to demonstrate the improvement it can achieve. To ensure a fair comparison, these networks were trained and tested on the SBCIV dataset using the same data augmentation methods. The training and testing of these models strictly followed the hyperparameter settings in their papers.
\end{itemize}

\subsection{Evaluation metrics}

To ensure a holistic and comparative analysis of the models' performance, metrics for assessing the segmentation performance are defined at the class-, task-, and model-levels. The class-level comparison focuses on the model's performance in handling the specific categories within each task. The task-level comparison evaluates the multi-task model's accuracy in performing individual tasks. The model-level comparison assesses the overall performance of the multi-task model.

For any of the datasets, a vector of binary variables, $\pmb{y}_{i,j}$, denotes the ground truth of class $j$ in task $i$ for all pixels in that dataset, and the other vector of binary variables, $\widehat{\pmb{y}}_{i,j}$, is the prediction. $\Lambda_i$ designates the set of classes in task $i$, for $i\in\Omega=\{\text{e}, \text{d}\}$. Here, $\Lambda_{\rm e}=$ \{Bearing, Bracing, Deck, Floor beam, Girder, Substructure, Background\} is the set of classes in the structural element segmentation task, and $\Lambda_{\rm d}=\{$Corrosion, Non-corrosion$\}$ is the set of classes in the defect segmentation task. 

In performing task $i$, a model's ability to predict class $j$ pixels is assessed using three widely recognized class-level metrics, namely Intersection over Union (IoU), Precision, and Recall:
\begin{equation}
\text{IoU}_{i,j}=\frac{\|\pmb{y}_{i,j}\wedge\widehat{\pmb{y}}_{i,j}\|_1}{\|\pmb{y}_{i,j}\vee \widehat{\pmb{y}}_{i,j}\|_1}
\label{eq:IoUij}
\end{equation}
\begin{equation}
\text{{Precision}}_{i,j}=\frac{\|\pmb{y}_{i,j}\wedge\widehat{\pmb{y}}_{i,j}\|_1}{\|\widehat{\pmb{y}}_{i,j}\|_1}
\label{eq:Precisionij}
\end{equation}
\begin{equation}
\text{{Recall}}_{i,j}=\frac{\|\pmb{y}_{i,j}\wedge\widehat{\pmb{y}}_{i,j}\|_1}{\|\pmb{y}_{i,j}\|_1}
\label{eq:Recallij}
\end{equation}
where $\wedge$ is the element-wise AND operator, $\vee$ is the element-wise OR operator, and $\|\cdot\|_1$ is norm 1 that can count the non-zero elements of the vector. $\text{IoU}_{i,j}$ in Eq.~(\ref{eq:IoUij}) calculates the intersection of the class $j$ ground truth and the prediction over their union, for any class $j\in\Lambda_i$. $\text{Precision}_{i,j}$ in Eq.~(\ref{eq:Precisionij}) is the percentage of pixels predicted as class $j$ which are predicted correctly, and $\text{Recall}_{i,j}$ in Eq.~(\ref{eq:Recallij}) is the percentage of class $j$ pixels that are correctly predicted.

For the task-level evaluation, three commonly used metrics were adopted, which are mean IoU (mIoU), mean Accuracy (mAcc), and pixel Accuracy (pAcc). For any task $i$, mIoU is calculated by averaging the IoU values of all the classes,
\begin{equation}
    \text{mIoU}_i=\frac{1}{|\Lambda_i|}\sum_{j\in\Lambda_i}\text{IoU}_{i,j}
\end{equation}
mAcc is the mean of class-level Recall values, 
\begin{equation}
\text{mAcc}_i=\frac{1}{|\Lambda_i|}\sum_{j\in\Lambda_i} \text{Recall}_{i,j}
\end{equation}
and pAcc denotes the micro-level accuracy evaluated without regard to the classes,
\begin{equation}
\text{pAcc}_i=\frac{\sum_{j\in\Lambda_i}\|\pmb{y}_{i,j}\wedge\widehat{\pmb{y}}_{i,j}\|_1}{\sum_{j\in\Lambda_i}\|\pmb{y}_{i,j}\|_1}
\end{equation}

Finally, at the model-level, using the two single-task models as the baseline, the study measured the incremental of an MTL model's overall performance against the baseline by averaging the percent increase of every task-level metric:
\begin{equation}
    \Delta=\frac{1}{|\Omega|\cdot|\Omega_{\rm m}|}\sum_{i\in\Omega,l\in \Omega_{\rm m}}\delta_{i,l},\quad \forall i\in\Omega \text{ and } l\in \Omega_{\rm m}
\end{equation}
where $\delta_{i,l}$ is the percent increase of metric $l$ in task $i$, $\Omega=\{\text{e},\text{d}\}$ is the index set of tasks, $\Omega_{\rm m}=\{\rm mIoU, \rm mAcc, \rm pAcc\}$ denotes the index set of task-level performance metrics, and $|\cdot|$ means the size of an index set.


\section{Experiments and results}
\label{sec:exp}
Experiments are conducted to verify the effectiveness of the proposed AECIF-Net on the newly developed SBCIV dataset.

\subsection{Ablation study}

An ablation study was performed to thoroughly evaluate the effectiveness of the key components designed for AECIF-Net. The single-task models utilizing the HRNetV2-W48, along with variants of the AECIF-Net, were trained and pitted against the AECIF-Net. Results are summarized in Table~\ref{tab:abla} and discussed below.

\begin{table*}[htbp]
	\begin{center}
        \caption{Ablation study for the evaluation of AECIF-Net's key components.}\label{tab:abla}
	\begin{threeparttable}
        \begin{tabular}{l|r r r| r r r}
        \hline
        \multirow{2}{*}{Method}&\multicolumn{3}{c|}{Element segmentation}&\multicolumn{3}{c}{Defect Segmentation}\\
        \cline{2-7}
          & mIoU  &  mAcc  & pAcc  & mIoU  &  mAcc  & pAcc\\ \hline
        Single-task & 91.86 & 95.34 & 96.96 & 84.88 & 90.34 & 96.91 \\
        Naive MTL & 89.60 & 94.86 & 96.61 & 85.26 & 90.54 & 97.07 \\
        AECIF-Net without CF & 91.83 & 95.25 & 96.40 & 85.61 & 90.00 & 97.20\\
        AECIF-Net without TF & 91.62 & 95.16 & 97.23 & 86.72 & 90.63 & 97.42\\
        AECIF-Net & \bf{92.11} & \bf{95.56} & \bf{97.27} & \bf{87.16} & \bf{90.79} & \bf{97.54}  \\ \hline
	\end{tabular}
	\begin{tablenotes}
        \footnotesize
        \item \textit{Note}: The highest value for each performance metric among all networks is in bold;\\
        MTL = Multi-task learning; CF = Co-interactive fusion module; TF = Task-specific feature relearning subnets.
        \end{tablenotes}
        \end{threeparttable}
	\end{center}
\end{table*}

From Table \ref{tab:abla}, it can be observed that the Naive MTL model favors the defect segmentation task more than the element segmentation task, as compared to the single-task models. Although the mIoU value of defect segmentation in the Naive MTL model slightly improves by 0.38\%, the mIoU value of element segmentation significantly drops by 2.26\%. Therefore, the Naive MTL model exhibits an overall decline in performance compared to the single-task models. It indicates that performing two different tasks by directly utilizing the overall feature embedding from the shared deep feature extractor is not as effective as the task-specific single-task models.

After introducing the two task-specific feature relearning subnets, the Naive MTL model becomes the AECIF-Net without CF model in Table \ref{tab:abla}. The added subnets effectively enhance the performance of the AECIF-Net without CF model, resulting in a respective improvement of 2.23\% and 0.35\% in the mIoU values of the two tasks, as compared to the Naive MTL model. The observed improvement verifies the effectiveness of further-learned task-specific features from the overall deep feature for the downstream tasks.

The AECIF-Net without TF is obtained by adding the co-interactive fusion module to the Naive MTL model. Compared to the single-task models, the AECIF-Net without TF maintains equivalent performance levels on the task of element segmentation yet displays a significant improvement in the task of defect segmentation. The comparison demonstrates the benefit of incorporating spatial information of structural elements in the defect segmentation task, and vice versa.

Different from the Naive MTL model, the proposed AECIF-Net has both the task-specific relearning subnets and the co-interactive fusion module. AECIF-Net effectively addresses the limitation of the Naive MTL model, evidenced by increases of the task-level metrics for 0.25$\sim$2.51\%. AECIF-Net exceeds the performance of the two single-task models on all metrics for 0.22$\sim$2.28\%.

\subsection{Comparison to state-of-the-art models}
This study comprehensively compares the proposed AECIF-Net with existing models by employing both quantitative experiments and qualitative evaluation, followed by model complexity assessment.

\subsubsection{Quantitative comparisons}
The quantitative comparisons were conducted at both task-level and class-level to be comprehensive.

{\bf Comparison at the task-level.} This study compared the proposed AECIF-Net with existing models: MaDnet~\citep{hoskere2020madnet}, MT-HRNet~\citep{https://doi.org/10.1002/stc.3128}, and MTL-I~\citep{doi:10.1177/03611981231155418}. The following discussion is mainly based on mIoU values. Observations from the model comparisons using other metrics, such as mAcc and pAcc, are similar.

Table~\ref{tab:task} shows that MaDnet, using a single-scale limited-capacity network, achieves 74.85\% and 78.66\% mIoU on the element segmentation task and defect segmentation task, respectively. MT-HRNet, which uses the most lightweight HRNetV2-W18 as the feature extractor, performs slightly better on the two tasks. However, among all the models considered, MaDnet and MT-HRNet exhibit the least favorable performance, which could be attributed to the utilization of less powerful encoders. Naive MTL model effectively boosts the two tasks' mIoU values to 89.60\% and 85.26\% by utilizing the most robust version of HRNet, HRNetV2-W48, as the encoder, although it keeps the same architecture as MT-HRNet. 
Despite the MTL-I model utilizing the less robust HRNetV2-W32 encoder, it achieves comparable outcomes on both tasks to the Naive MTL model that employs the more powerful HRNetV2-W48 encoder. That is, the performance improvement that an MTL model can gain from integrating the feature projection and cross-talk feature sharing is approximately equivalent to the improvement obtained from substituting the encoder HRNetV2-W48 for HRNetV2-W32.
Among all the compared methods, AECIF-Net achieves the best performance in all the metrics, whose mIoU values are 17.26\% and 8.50\% higher than MaDnet's values, further verifying the advantage of the proposed approach.

\begin{table*}[htbp]
    \caption{Quantitative comparison results with state-of-the-art methods at the task-level, highlighting AECIF-Net's superior performance across all metrics.}\label{tab:task}
	\begin{center}
	\begin{threeparttable}
        \begin{tabular}{l|r r r| r r r}
        \hline
        \multirow{2}{*}{Method}&\multicolumn{3}{c|}{Element segmentation}&\multicolumn{3}{c}{Defect Segmentation}\\
        \cline{2-7}
          & mIoU  &  mAcc  & pAcc  & mIoU  &  mAcc  & pAcc\\ \hline
        MaDnet~\citep{hoskere2020madnet} & 74.85 & 83.54 & 90.66 & 78.66 & 84.58 & 95.67 \\
        MT-HRNet~\citep{https://doi.org/10.1002/stc.3128} & 78.09 & 87.88 & 91.56 & 79.76 & 85.39 & 95.93  \\
        Naive MTL & 89.60 & 94.86 & 96.61 & 85.26 & 90.54 & 97.07 \\
        MTL-I~\citep{doi:10.1177/03611981231155418} & 89.07 & 94.54 & 96.17 & 84.30 & 90.07 & 96.84  \\
        \textbf{AECIF-Net (Ours)} & \textbf{92.11} & \textbf{95.56} & \textbf{97.27} & \textbf{87.16} & \textbf{90.79} & \textbf{97.54}  \\ \hline
		\end{tabular}
		\begin{tablenotes}
        \footnotesize
        \item \textit{Note}: The highest value for each performance metric among all networks is in bold.
        \end{tablenotes}
    \end{threeparttable}
	\end{center}
\end{table*}

\begin{table*}[tb]
        \caption{Class-level IoU metric comparison showing AECIF-Net's remarkable performance against state-of-the-art methods across all classes.}\label{tab:iou}
	\begin{center}
        \begin{threeparttable}
        \begin{tabular}{l|r r r r r}
        \hline
        \multirow{2}{*}{\diagbox{Class}{Method}}&\multicolumn{1}{l}{MaD-}&\multicolumn{1}{l}{MT-}&\multicolumn{1}{l}{Naïve}&\multicolumn{1}{l}{MTL}&\multicolumn{1}{l}{\textbf{AECIF-Net}}\\
        \multicolumn{1}{l}{}&\multicolumn{1}{|l}{net~\citep{hoskere2020madnet}}&
                           \multicolumn{1}{l}{HRNet~\citep{https://doi.org/10.1002/stc.3128}}&
                           \multicolumn{1}{l}{MTL}&
                           \multicolumn{1}{l}{-I~\citep{doi:10.1177/03611981231155418}}&
                           \multicolumn{1}{r}{\textbf{(Ours)}}\\ \hline
        Element &  & &  &  &  \\
        \quad Bearing & 72.69 & 74.76 & 90.47 & 88.32 & \textbf{91.68}\\
        \quad Bracing & 59.07 & 66.00 & 86.63 & 86.81 & \textbf{92.11} \\
        \quad Deck & 79.81 & 87.76 & 92.77 & 90.99 & \textbf{94.89} \\
        \quad Floor beam & 68.30 & 67.56 & 79.53 & 85.82 & \textbf{86.19} \\
		\quad Girder & 88.81 & 88.76 & 95.61 & 95.65 & \textbf{96.68} \\
		\quad Substructure & 87.91 & 89.39 & 96.02 & 94.35 & \textbf{96.16} \\
		\quad Background & 67.38 & 72.49 & 86.21 & 81.58 & \textbf{87.09} \\
		Defect&  & & & & \\
		\quad No Corrosion & 95.34 & 95.61 & 96.80 & 96.56 & \textbf{97.32}\\
		\quad Corrosion & 61.97 & 63.90 & 73.71 & 72.05 & \textbf{76.99}\\\hline
		\end{tabular}
		\begin{tablenotes}
        \footnotesize
        \item \textit{Note}: The highest values of each metric for each class among all networks are shown in bold.
        \end{tablenotes}
        \end{threeparttable}
	\end{center}
\end{table*}

{\bf Comparison at the class-level.} AECIF-Net also achieves good performance at the class-level compared to state-of-the-art models, as shown in Table \ref{tab:iou}. AECIF-Net clearly outperforms the competitors among all classes. MaDnet and MT-HRNet are notably less capable of segmenting bracings and floor beams than other types of elements, which could be attributed to their irregular shapes. For example, bracings are predominantly cross-shaped. Compared to MaDnet, the Naive MTL model increases the IoU values in segmenting bracings by 27.56\% and 11.23\% for segmenting floor beams. The significant improvement is merely due to upgrading the encoder to a more capable one. Compared to the Naive MTL model, the MTL-I model further increases the IoU in segmenting bracings for another 0.18\% and 6.29\% in segmenting floor beams, indicating the effectiveness of feature disentanglement and information sharing in segmenting irregularly shaped elements. Ultimately, AECIF-Net achieves the highest IoUs, 92.11\% in segmenting bracings and 86.19\% in segmenting floor beams. The increased IoUs for another 5.30\% and 0.37\% that AECIF-Net achieved, compared to MTL-I, indicate that the relearning subnets and the co-interactive fusion module are better designs than their counterparts in MTL-I. In segmenting girders and substructures, which represent the majority of pixels across all classes, all models demonstrate at least acceptable results. A noticeable trend is observed, where the IoU value progressively increases from the leftmost to the rightmost model. The progress improvements are mainly due to the introduction of a powerful deep feature extractor, feature disentanglement or relearning, and feature fusion to MTL. The increases in IoU values in segmenting bearings and decks are primarily attributed to using HRNet as the deep feature extractor.

AECIF-Net also demonstrates dominantly better performance in segmenting defects than other models. Compared to MaDnet, AECIF-Net increases the IoU value in segmenting corrosion by 15.02\%, with 11.74\% contributed by the adoption of HRNetV2-W48 as the backbone and the remaining 3.28\% from feature relearning and co-interactive feature fusion. AECIF-Net also increases the ability to segment non-corrosion areas by 1.98\%.

\subsubsection{Qualitative comparison}

\begin{figure}[htbp]
    \centering
    \includegraphics[width=\textwidth]{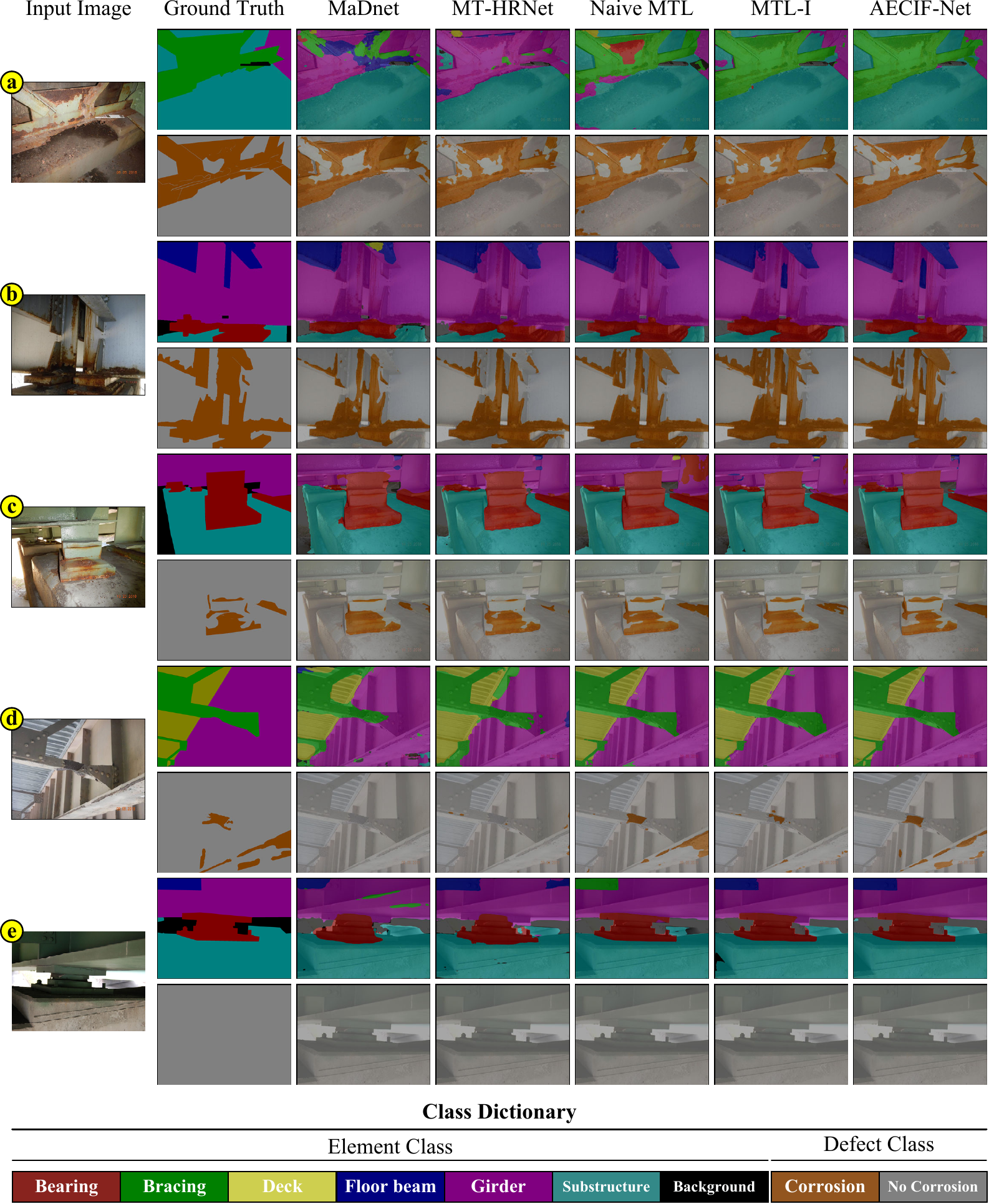}
    \caption{Qualitative comparison with state-of-the-art, where AECIF-Net excels in segmenting small or irregular elements, as seen in (a), (b), (c), (d), and (e), and it offers more precise edges, evident in (b) and (e); AECIF-Net produces fewer false positives in defect segmentation, especially in small corrosion areas observed in (b), (c), and (d).}
    \label{fig:sota}
\end{figure}

Fig. \ref{fig:sota} illustrates five examples of bridge element segmentation and defect segmentation results to demonstrate the effectiveness of the proposed AECIF-Net. These examples represent various scenarios with extensive, partial, and scarcely defects.

The qualitative evaluation of the element segmentation results demonstrates that AECIF-Net generates superior predictions, as evidenced by the segmentation of small objects, such as the distant floor beam in Fig.~\ref{fig:sota}(b), bearings in Fig.~\ref{fig:sota}(c), distant small bracing in Fig.~\ref{fig:sota}(d), and floor beam in Fig.~\ref{fig:sota}(e). This is also apparent in segmenting irregular elements, as demonstrated by the bracings in Fig.~\ref{fig:sota}(a)(d). Furthermore, AECIF-Net facilitates more pronounced and smooth edges of object segmentation, as evidenced by the boundary of floor beam segmentation in Fig.~\ref{fig:sota}(b) and substructure in Fig.~\ref{fig:sota}(e).

The qualitative comparison of the defect segmentation results reveals that AECIF-Net can produce competitive outcomes, as it generates fewer false positives than other methods when the corrosion area is small or less visible, which is evident in Fig.~\ref{fig:sota}(b)(c)(d).

\begin{figure*}[htbp]
    \centering
    \includegraphics[width=\textwidth]{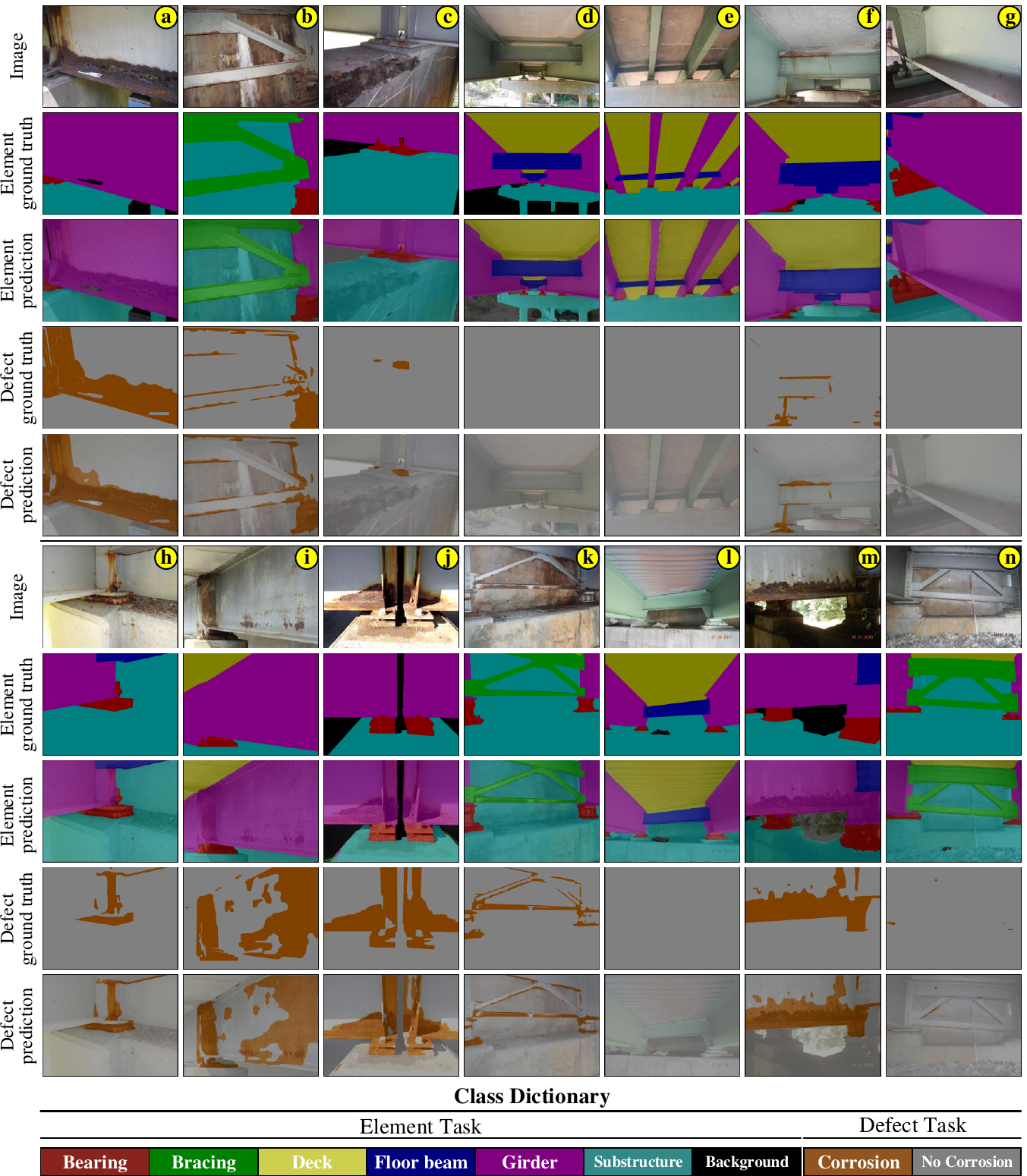}
    \caption{Examples of AECIF-Net on the testing set demonstrate its high-quality predictions in close alignment with the ground truth.}
    \label{fig:test}
\end{figure*}

Figure \ref{fig:test} further presents AECIF-Net's results in analyzing fourteen examples from the testing set. It is evident that the predictions given by AECIF-Net exhibit high quality and closely align with the ground truth.

\subsubsection{Model complexity comparison}
 
\begin{figure*}[htbp]
    \centering
    \includegraphics[width=0.7\textwidth]{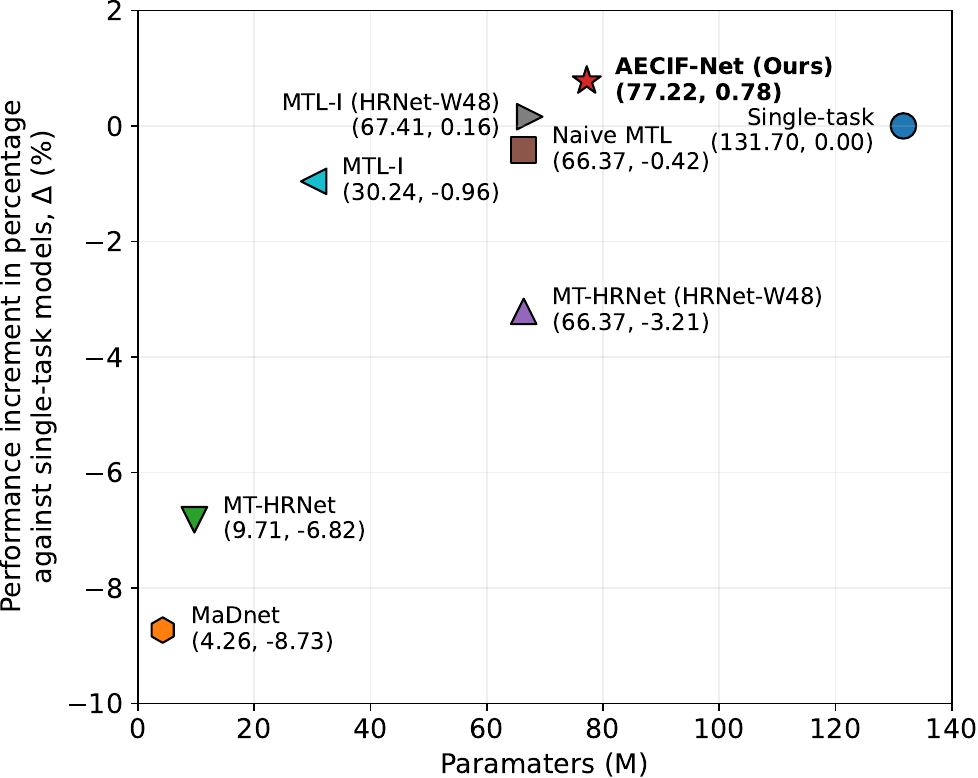}
    \caption{Comparisons of models by their performance and complexity, which use single-task models as the baseline for comparison. The comparison shows that AECIF-Net achieves the best balance between performance and complexity among all models.}
    \label{fig:source}
\end{figure*}

A deep learning model's complexity is a cost for the model's performance improvement.
Therefore, the performance assessment should keep the model complexity in consideration.
The count of trainable parameters in a DCNN is a straightforward and widely recognized metric for quantifying model complexity. In this study, it demonstrates a high degree of correlation with both calculated floating point operations (FLOPs) and inference time. Consequently, the count of trainable parameters is used to measure the model's complexity.
Figure \ref{fig:source} presents AECIF-Net and other state-of-the-art models on the diagram of model-level performance increment, $\Delta$, in percentage vs. model complexity (parameters in millions).

MaDnet has the fewest parameters due to its simplistic structure, while MT-HRNet has slightly more parameters because it utilizes the lightweight HRNetV2-W18. Consequently, the performance of these two models is less than ideal. MTL-I, which employs HRNetV2-W32 as its backbone, results in a significant performance improvement compared to MaDnet and MT-HRNet. The performance of MTL-I still falls below the single-task baseline, but its parameters are about 77\% less than the baseline. By replacing the backbone of MTL-I with HRNetV2-W48, the upgraded version, MTL-I (HRNetV2-W48), outperforms its HRNetV2-W32 counterpart and the single-task models. When MT-HRNet's backbone is changed to HRNetV2-W48, the newer version, MT-HRNet (HRNetV2-W48), and Naive MTL share the same architecture and the number of parameters, yet Naive MTL performs substantially better due to the optimized selection of hyperparameters. 
AECIF-Net achieves the best performance among all models, although the number of parameters it utilizes is only about 60\% of single-task models' parameters. The single-task models are a collection of two independent networks, with each dedicated to one task. Each independent network has 65.85 million parameters, leading to a total of 131.70 million parameters. In contrast, AECIF-Net streamlines this with a unified network of 77.22 million parameters, a reduction primarily due to its shared encoder. Notably, the inclusion of AECIF-Net's task-specific subnets (3.32$\times$2=6.64 million) and the co-interactive fusion module (2.36$\times$2=4.72 million) adds another 11.36 million parameters. Converting features with different dimensions ($720\times120\times120$ in single-task models and $512\times120\times120$ in AECIF-Net) into segmentation predictions also needs distinct parameter quantities for the reconstruction decoders (0.52$\times$2=1.04 million for single-task models while 0.267$\times$2$\approx$0.53 million for AECIF-Net). Consequently, AECIF-Net achieves a net saving of 54.48 ($=65.33-11.36-0.53+1.04$) million parameters, underscoring its efficiency in its overall reduced parameter count. Compared to MaDnet, AECIF-Net's complexity has increased by 72.96 million, with the majority (61.60 million) added by replacing the original backbone with HRNetV2-W48 for representation learning and the remainder (11.36 million) from the relearning subnets and co-interactive fusion module.

\subsection{Understanding of feature maps and masks}

\begin{figure*}[htbp]
    \centering
    \includegraphics[width=\textwidth]{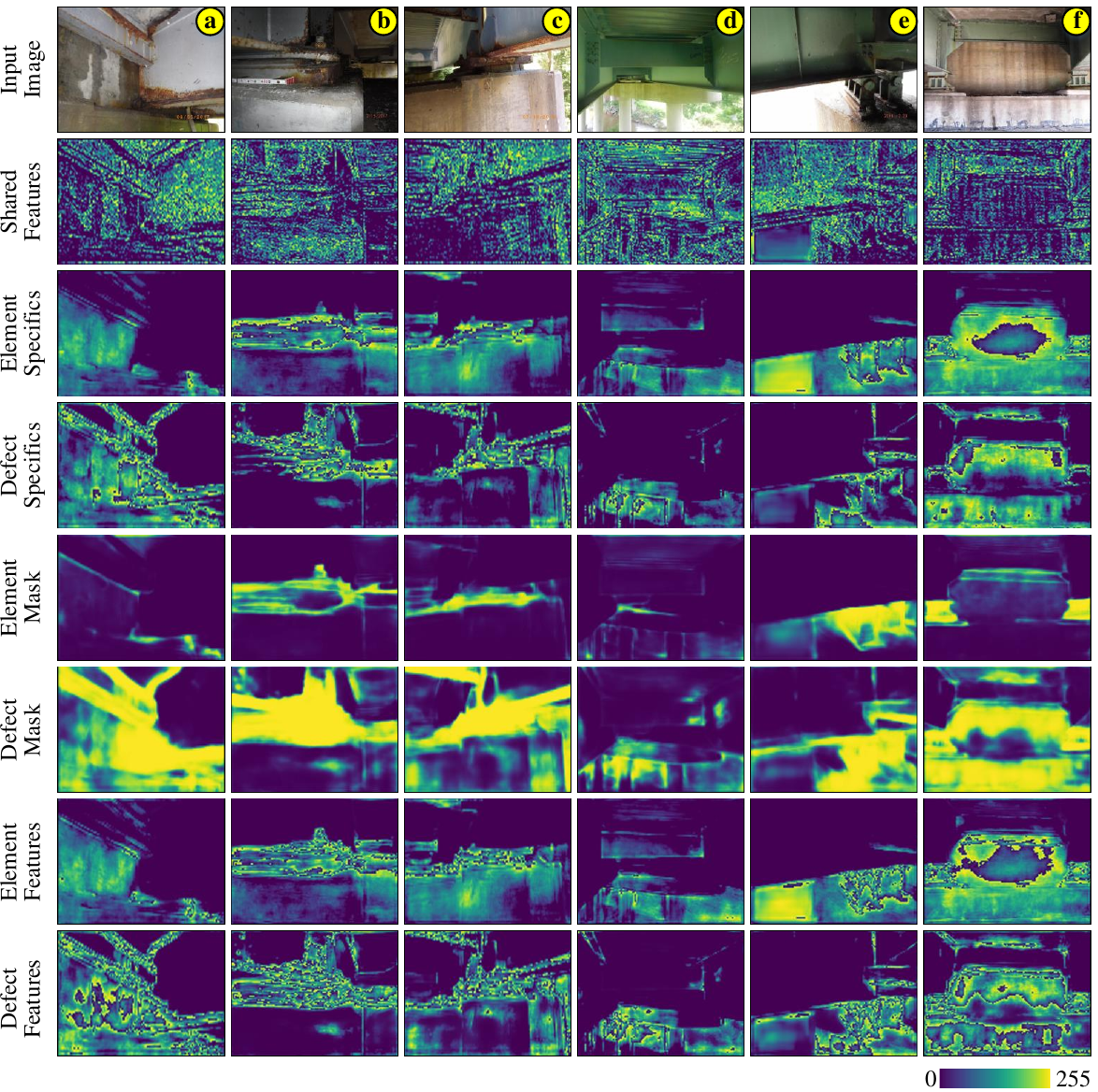}
    \caption{The evolution of image features from the overall embedding to task-specific features, along with the attention-enhanced co-interactive feature fusion in AECIF-Net, reveals the reasons behind its effective multi-task performance.}
    \label{fig:mask}
\end{figure*}

Task-specific feature relearning and co-interactive feature fusion are two important designs for AECIF-Net. To better comprehend the roles of those modules, feature maps ($\pmb{f}$, $\pmb{f}_{\rm e}$, $\pmb{f}_{\rm d}$, $\pmb{f}_{\rm e}^\ast$, $\pmb{f}_{\rm d}^\ast$) and attention masks ($\pmb{S}_{\rm e}$ and $\pmb{S}_{\rm d}$) learned by AECIF-Net are visualized for six examples in Fig.~\ref{fig:mask}. 
These visualizations utilize the feature embeddings from the second channel to show primary features, with the values of each image rescaled to be within the range from 0 to 255 to accommodate the colors. Bicubic interpolation is applied to resize the feature maps and attention masks to $520\times 520$, the dimension of the input images of AECIF-Net.

The 2nd row in Fig.~\ref{fig:mask} visualizes the overall feature embedding $\pmb{f}$ that essentially encompasses information relevant to both tasks. With the two feature relearning subnets, the element-specific feature map $\pmb{f}_{\rm e}$ and defect-specific feature map $\pmb{f}_{\rm d}$ are respectively extracted from the overall feature embedding. The 3rd row visualizes the element-specific feature maps that primarily capture relatively global information about elements, such as position, shape, and scale. In contrast, the defect-specific feature maps visualized in the 4th row mainly capture appearance information of surface defects, such as texture and color. The 5th row visualizes the element masks $\pmb{S}_{\rm e}$, whereas the 6th row presents the defect masks $\pmb{S}_{\rm d}$. A distinct difference in the two tasks' attention masks is evident. Each task's mask functions as a feature selector that masks out uninformative portions of the other task's feature map in feature fusion. It is noteworthy that the element masks exhibit a significantly higher contrast, transitioning from blue to yellow, while the defect masks predominantly present a yellow hue in the same area. This finding implies that the element task benefits more from the extraction of task-specific features. The 7th and 8th rows are spatial-attention-enhanced feature maps $\pmb{f}_{\rm e}^\ast$ and $\pmb{f}_{\rm d}^\ast$. Compared to $\pmb{f}_{\rm e}$ in the 3rd row, $\pmb{f}_{\rm e}^\ast$ in the 7th row captures more detailed appearance information about the elements and thus enhances the ability to differentiate different types of elements. Similarly, compared to $\pmb{f}_{\rm d}$ (the 4th row),  $\pmb{f}_{\rm d}^\ast$ (the 8th row) integrates the context information about where the defect is developed, which enhances the ability to differentiate defects from defect-like texture such as watermarks on substructures and shadows.


\section{Case study on structural condition assessment}
\label{sec:case}

\begin{figure*}[htbp]
    \centering
    \includegraphics[width=\textwidth]{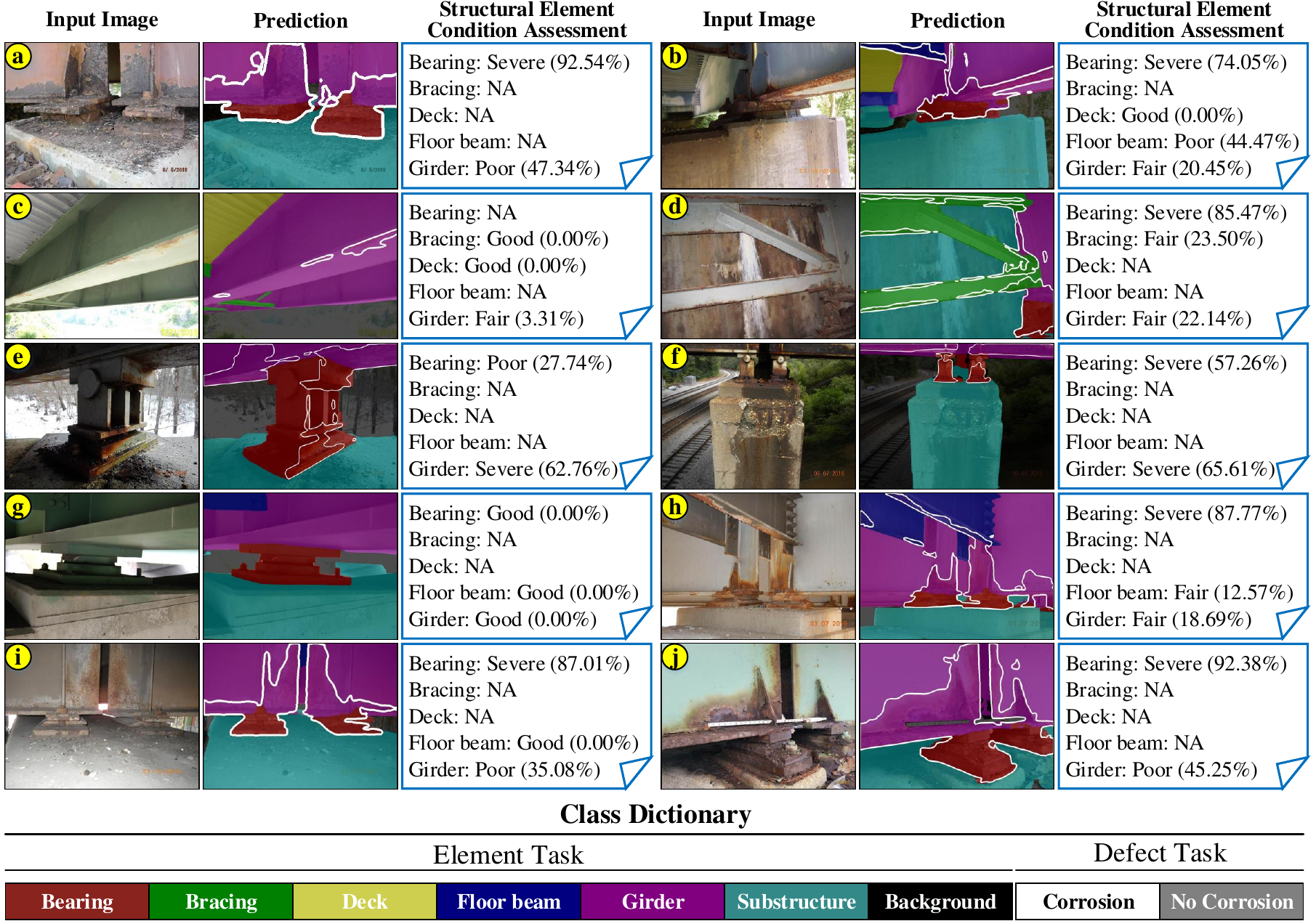}
    \caption{Structural condition assessment using AECIF-Net showcases its ability to accurately assess the conditions of bridge elements, underlining its potential in infrastructure inspection.}
    \label{fig:assess}
\end{figure*}

The potential application of AECIF-Net in bridge inspection is further illustrated in a preliminary structural condition assessment case study. To illustrate how AECIF-Net performs the visual assessment for bridge elements, ten examples in the testing set are presented in Fig.~\ref{fig:assess}. Each steel element is preliminarily evaluated based on the proportion of the corroded area observed on the element (the ratio of the corroded area to the entire element area). The structural conditions of bridge elements are then categorized into four distinct levels: Good, Fair, Poor, and Severe. These classifications are determined by specific thresholds of corrosion coverage: 0\%, 25\%, and 50\%. These thresholds, which define the intervals between the various condition classifications, are graphically depicted in Fig.~\ref{fig:fram}.

AECIF-Net demonstrates promising initial evaluation results in scenarios where the bridge elements are in good condition, as illustrated in Fig.~\ref{fig:assess}(c)(g). Moreover, the proposed method is capable of accurately identifying the elements of interest in cases with severe damage, even when there is extensive surface deterioration. The AECIF-Net provides precise preliminary evaluation outcomes in such instances, as displayed in Fig.~\ref{fig:assess}(a)(d)(e)(f)(j). In scenes featuring a mixture of elements in both well or unsatisfactory conditions, AECIF-Net is capable of accurately determining the condition of each element, which can be observed in Fig.~\ref{fig:assess}(b)(h)(i). This level of accuracy and reliability in detecting and assessing both well-maintained and heavily damaged bridge elements underscores the potential of AECIF-Net as a valuable tool in the field of infrastructure inspection.

\section{Conclusions}
\label{sec:con}

This paper presented a deep learning architecture called Attention-Enhanced Co-Interactive Fusion Network (AECIF-Net) and a newly annotated Steel Bridge Condition Inspection Visual (SBCIV) dataset, to automate visual bridge inspection for structural condition assessment. AECIF-Net, employing a share-split-interaction pipeline, is a unified model that can simultaneously segment bridge structural elements and surface defects on them from inspection images. The two tasks share a deep, high-resolution encoder HRNetV2-W48, but they have their own task-specific relearning subnets. The co-interactive feature fusion module further improves the task-specific features with spatial-attention-enhanced feature fusion. AECIF-Net has surpassed the current state-of-the-art MTL methods for structural condition assessment, achieving 92.11\% element segmentation mIoU and 87.16\% corrosion segmentation mIoU on the test set of SBCIV. The ablation study revealed how AECIF-Net's key modules contribute to performance improvement, and the evaluation on the test set further demonstrated the capability of AECIF-Net in automated visual inspection.

While AECIF-Net has shown promising results in extracting structural elements in inspection images and assessing the elements' conditions, addressing limitations presented in the current work will broaden the impact of AECIF-Net on the visual inspection of civil infrastructure. One obstacle to be addressed is the scarcity of annotated data. Bridges are diverse in types and designs. The condition of the same bridge is also changing over time due to deterioration or reparation. The performance of AECIF-Net will drop by a certain amount if the inspection images contain new elements or new defect types. Annotated data are required to adapt AECIF-Net to new tasks. An efficient data annotation tool is desired to accommodate the need for annotated data.
Besides, in drone-assisted bridge inspections, challenges like varying image quality, inconsistent capture angles, motion blurs, and reflections can impact the effective training of developed models. To counter these issues, preprocessing techniques such as contrast adjustment, geometric modifications, deblurring algorithms, and color corrections can be employed for better adaptability under challenging data acquisition scenarios. Furthermore, AECIF-Net's enhanced task performance is partly attributed to incorporating a deep feature encoder.
Deep feature extraction is the most time-consuming portion of image analysis. Lightweight feature extractors that are as powerful as the deep feature extractors are desired because the runtime efficiency would support inspectors' decisions at inspection fields, such as utilizing additional advanced inspection methods (e.g., infrared cameras, ground-penetrating radar, ultrasound scanning) to collect data at concerned areas identified from the visual inspection. Those future studies will move the research on this topic forward, and the automation of infrastructure inspection will continue blooming.


\section{Acknowledgement}
\label{sec:sample:appendix}

The authors of this work receive support from the National Science Foundation (grant numbers ECCS-2025929, ECCS-2026357). Any opinions, findings, conclusions, or recommendations expressed in this material are those of the authors and do not necessarily reflect the views of the National Science Foundation.

\bibliographystyle{elsarticle-num-names} 
\bibliography{cas-refs}





\end{document}